\documentclass[10pt,twocolumn,letterpaper]{article}

\usepackage{cvpr}
\usepackage{times}
\usepackage{epsfig}
\usepackage{graphicx}
\usepackage{amsmath}
\usepackage{amssymb}

\usepackage[bookmarks=false,pagebackref=true,breaklinks=true,letterpaper=true,colorlinks]{hyperref}

\usepackage[square,numbers,sort]{natbib}
\usepackage{algpseudocode}
\usepackage{booktabs}
\usepackage{array}
\usepackage{multirow}
\usepackage{algorithm}
\usepackage{caption}
\usepackage{subcaption}
\usepackage[normalem]{ulem}

\newcommand{\approptoinn}[2]{\mathrel{\vcenter{
  \offinterlineskip\halign{\hfil$##$\cr
    #1\propto\cr\noalign{\kern2pt}#1\sim\cr\noalign{\kern-2pt}}}}}

\cvprfinalcopy 

\def\cvprPaperID{542} 

\ifcvprfinal\fi
\begin{document}

\title{Watch and Learn: Semi-Supervised Learning of Object Detectors from Videos} 

\author{
    Ishan Misra \quad \quad
    Abhinav Shrivastava \quad \quad
    Martial Hebert \vspace{0.05in} \\
    Robotics Institute, Carnegie Mellon University \\
    \footnotesize{\texttt{\{imisra,ashrivas,hebert\}@cs.cmu.edu }} \\
}

\maketitle

\begin{abstract}
We present a semi-supervised approach that localizes multiple unknown object instances in long videos. We start with a handful of labeled boxes and iteratively learn and label hundreds of thousands of object instances. We propose criteria for reliable object detection and tracking for constraining the semi-supervised learning process and minimizing semantic drift. Our approach does not assume exhaustive labeling of each object instance in any single frame, or any explicit annotation of negative data. Working in such a generic setting allow us to tackle multiple object instances in video, many of which are static. In contrast, existing approaches either do not consider multiple object instances per video, or rely heavily on the motion of the objects present. The experiments demonstrate the effectiveness of our approach by evaluating the automatically labeled data on a variety of metrics like quality, coverage (recall), diversity, and relevance to training an object detector.
\vspace{-0.2in}
\end{abstract}

\section{Introduction}

The availability of large labeled image datasets~\cite{pascal-voc-2007,imagenet} has been one of the key factors for advances in
recognition. These datasets, which have been largely curated from internet
images, have not only helped boost performance~\cite{pedro-dpm,rcnn}, but have
also fostered the development of new techniques~\cite{alexnet,rcnn}. However, compared to images, videos seem like a more natural source of training data because of the additional temporal continuity they offer for both learning and labeling. So ideally we should have large labeled internet video datasets. In general, the human effort required for labeling these vision datasets is huge, e.g., ImageNet~\cite{imagenet} required 19 man-years to label bounding boxes in the 1.2 million images harvested from the
internet. Consider the scale of internet videos -- YouTube
has 100 hours of video (10 million frames) uploaded every minute. It seems
unlikely that human per-image labeling will scale to this amount of data.
Given this scale of data and the associated annotation problems~\cite{labelme-video,virat}, which are more pronounced in videos,
it is no surprise that richly annotated large video recognition datasets are
hard to find. In fact, the available video datasets~\cite{ucf101,hollywood,sports1m,virat} lack the kind
of annotations offered by benchmark image datasets~\cite{pascal-voc-2007,coco,imagenet}.
\begin{figure}[!t]
\centering
\includegraphics[width=0.45\textwidth]{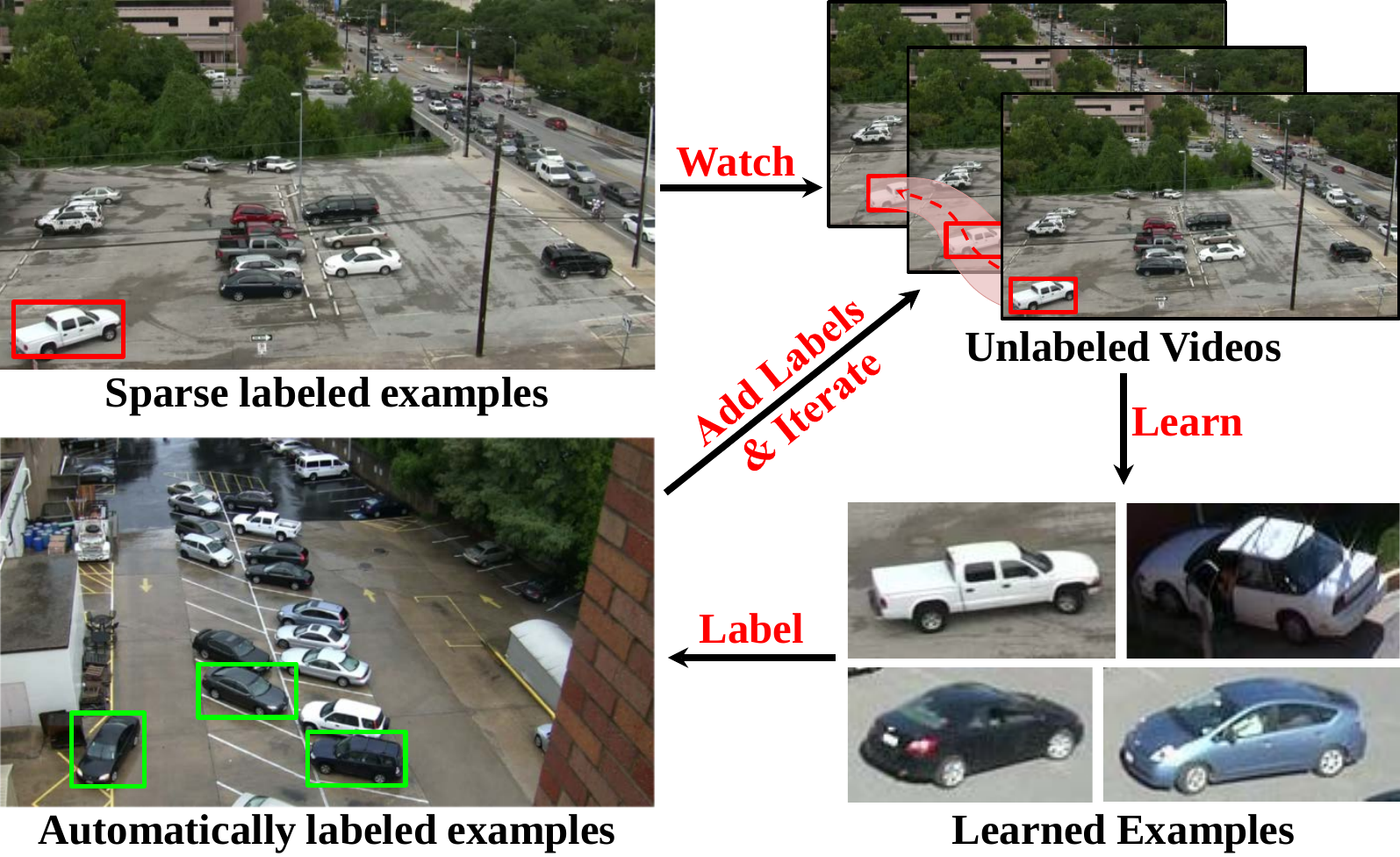}        
\vspace{-0.1in}
\caption{We present a novel formulation of semi-supervised learning for automatically learning object detectors from videos. Our method works with long video to automatically learn bounding box level annotations for multiple object instances. It does not assume exhaustive labeling of every object instance in the input videos, and from a handful of labeled instances can automatically label hundreds of thousands of instances.}
\vspace{-0.2in}
\label{fig:pullup}
\end{figure}

One way to tackle the labeling problem is using semi-supervised learning (SSL).
Starting with only a few annotated examples, the algorithm can label more
examples automatically. However, a major challenge for any kind of SSL technique is to
constrain the learning process to avoid semantic drift, i.e., added 
noisy samples cause the learner to drift away from the true concept. Recent
work~\cite{abhinav-ssl,neil,lean,choi-ssl} has shown ways to constrain this learning
process for images. In this paper, we present an approach to constrain the semi-supervised learning process~\cite{abhinav-ssl} in videos. Our technique
constrains the SSL process by using \emph{multiple weak cues} - appearance,
motion, temporal etc., in video data and automatically learns \emph{diverse
new examples}.

Intuitively, algorithms dealing with videos should use appearance and temporal
cues using detection and tracking, respectively. One would expect a
simple combination of detection and tracking to constitute a semi-supervised
framework that would prevent drift since both of these processes
would ideally cancel each others' errors. However, as we show in our
experiments (Sec.~\ref{sec:experiments}), a na\"{\i}ve combination of these two
techniques performs poorly. In the long run, the errors in both detection and
tracking are amplified in a coupled system. We can also consider pure
detection approaches or pure tracking approaches for this problem.
However, pure detection ignores temporal information while pure tracking tends
to stray away over a long duration.

\begin{figure*}[!ht]
\centering
\includegraphics[width=0.9\textwidth]{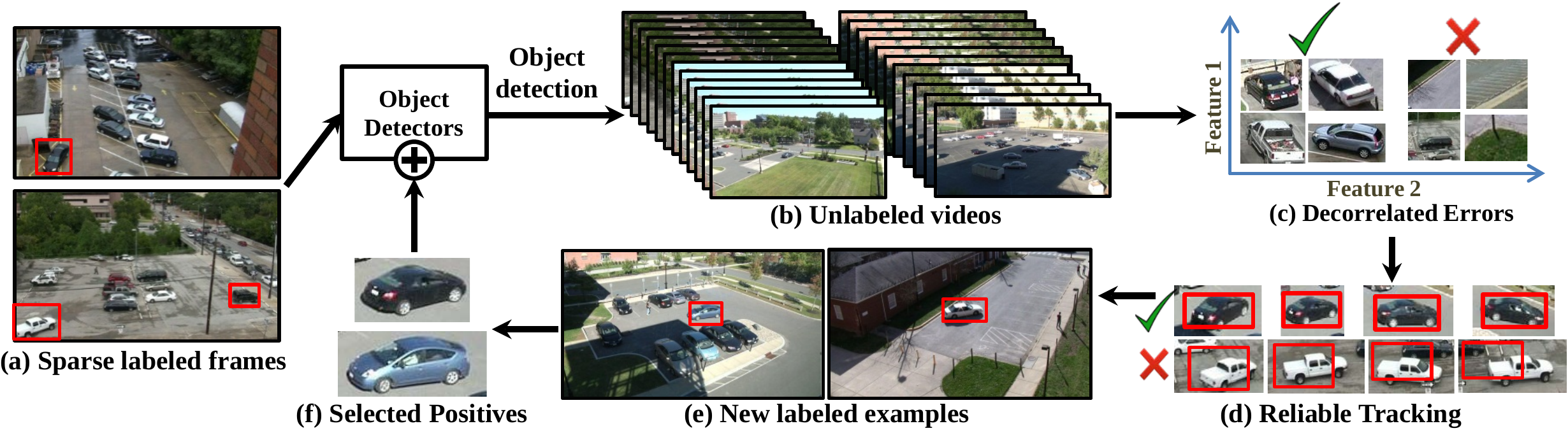}        
\vspace{-0.1in}
\caption{Our approach selects samples by iteratively discovering new boxes by a careful fusion of detection, robust tracking, relocalization and multi-view modeling of positive data. It shows how an interplay between these techniques can be used to learn from large scale unlabeled video corpora.}
\vspace{-0.2in}
\label{fig:teaser}
\end{figure*}

We present a scalable framework that discovers objects in
video using SSL (see Figure~\ref{fig:pullup}). It tackles the challenging problem of localizing new object instances in long videos starting from
only a few labeled examples. In addition, we present our algorithm in a
realistic setting of ``sparse labels''~\cite{virat}, i.e., in the few initial ``labeled''
frames, not all objects are annotated. This setting relaxes the assumption that in a given frame, all object instances have been exhaustively annotated. It implies that we do not know if any unannotated region in the frame belongs to the object category or the background, and thus cannot use any region from our input as negative data.
While much of the past work has ignored this type of sparse labeling (and \emph{lack of explicit negatives}), we show ways to overcome this handicap. Figure~\ref{fig:teaser} presents an overview of our algorithm.
Our proposed algorithm is different from the rich body of work on tracking-by-detection. Firstly, we do not attempt to solve the
problem of \emph{data association}. Our framework does not try to identify
whether it has seen a particular instance before. Secondly, since it works in
the regime of \emph{sparse annotations}, it does not assume that negative data
can be sampled from around the current box. Thirdly, we limit expensive
computation to a subset of the input frames to scale to a million frames.

\par \noindent \textbf{Contributions:} We present a semi-supervised learning framework that
\emph{localizes multiple unknown objects} in videos. Starting from few
\emph{sparsely labeled} objects, it iteratively labels new and useful training examples in the videos. Our key contributions are: 1) We tackle the SSL problem for discovering multiple objects in sparsely labeled videos; 2) We
present an approach to constrain SSL by combining multiple weak cues in
videos and exploiting decorrelated errors by modeling data in multiple feature spaces.
We demonstrate its effectiveness as compared to traditional tracking-by-detection approaches; 3) Given the redundancy in video data, we need a
method that can automatically determine the relevance of training examples to
the target detection task. We present a way to include \emph{relevance and diversity of the training examples}
in each iteration of the SSL process, leading to a scalable \emph{incremental learning} algorithm.

\vspace{-0.1in}
\section{Related Work}
\vspace{-0.05in}
\label{sec:relatedWork}

The availability of web scale image and video data has made semi-supervised learning more popular in recent years. In the case of images, many methods~\cite{fergus-ssl,ensemble-projection} rely on image similarity measures, and try to assign similar labels to close-by unlabeled images. However, in the case of real-world images, obtaining good image similarity is hard and hence the simple approaches become less applicable. One major body of work~\cite{abhinav-ssl,neil,lean,choi-ssl,devi-ssl} tries to overcome this difficulty by leveraging the use of a set of pre-defined attributes for image classes~\cite{abhinav-ssl,choi-ssl} and additionally web-supervision and text~\cite{lean,neil}. While these methods have good performance for images, they are not applicable to videos mainly because they treat each image independently and do not use video constraints. One major reason for the success of attribute based methods for SSL was the relatively cheap supervision required for attributes (per image level tag). In the same spirit, weakly supervised video approaches use tags available with internet videos.

Weakly supervised video algorithms have gained popularity largely due to the abundance of video level tags on the internet. The input is a set of videos with video level tags (generally a video belongs to an object category), and the algorithm discovers the object (if present) in the video. These methods, while effective, assume a maximum of one dominant object per video~\cite{prest-video,wang-discov,tang-segment,liu-vid-discov}. Some of them additionally assume dominant motion~\cite{prest-video} or appearance saliency~\cite{prest-video,wang-discov,liu-vid-discov,tang-segment,yongjae-egocentric} for the object of interest. The methods of video co-segmentation~\cite{chen-video-coseg,jose-video-coseg,lin-video-coseg,guo-video-coseg} can be considered a subset of weakly supervised methods. They make a strong assumption that multiple videos contain the exact same object in majority of the frames.
This assumption of at most one salient object in a video is rarely satisfied by internet or real world videos. When this assumption does not hold, methods cannot strongly rely on motion based foreground/background clustering or on appearance saliency. Our proposed work deals with \emph{multiple objects} and can even discover \emph{static} object instances without strongly relying on motion/appearance saliency. However, we do require richer bounding box annotations by way of a few \emph{sparsely labeled} instances. A concurrent work~\cite{babylearning} utilizes weakly labeled video for improving detection.

A relevant thread of work which also uses bounding box annotations is that of tracking-by-detection. It has a long and rich history in computer vision and the reader is referred to~\cite{tracking-benchmark} for a survey. The tracking-by-detection algorithms start with bounding box annotation(s) of the object(s) to track the object(s) over a long period of time. The underlying assumption is that negative data can be sampled from around the object~\cite{tld,spltt,ensemble-tracking,struck,mvboost-track,sslboost-track,thrun-ssltracking} to distinguish between the object and background. This is not valid in the case of \emph{sparsely labeled} videos because the unmarked regions may contain more instances of the same object, rather than background.

Other tracking-by-detection methods~\cite{hamed-deva-tracking,k-shortest,geiger-kitti-2} do not sample negative data from the input video. They do so at the cost of using a detector trained on additional training data, e.g., ~\cite{hamed-deva-tracking} uses a DPM~\cite{pedro-dpm} trained on images from PASCAL~\cite{pascal-voc-2007}. In contrast, we do not use such additional training data for our method. This allows us to work on object categories which are not in standard datasets.

Multi-object tracking-by-detection approaches also focus on solving the problem of \emph{data association} - given object locations across multiple frames, determine which locations belong to the same object. Data association is critical for long term tracking, and is also very challenging~\cite{k-shortest}. In contrast, our goal is not to track over long periods, but to get short and reliable tracking. Moreover, we do not require these short tracklets to be associated with each other, and thus have minimal need for data association.

To summarize, our SSL framework operates in a less restrictive domain compared to existing work in weakly labeled object discovery and tracking-by-detection. The key differences are: 1) We localize multiple objects in a single frame as opposed to zero or one objects. 2) We do not assume strong motion or appearance saliency of objects, thus discovering static object instances as well. 3) We operate in the regime of \emph{sparsely labeled} videos. Thus, in any given frame, all the unmarked region may contain instances of the object. This does not allow using negative data from the input frame. 4) We \emph{do not need explicit negative data} or any pre-trained object models. 5) Finally, the aim of our approach is very different from tracking approaches. We do not want to track objects over a long period of time, but want short reliable tracklets.

\begin{figure}[t]
        \centering
        \includegraphics[width=0.48\textwidth]{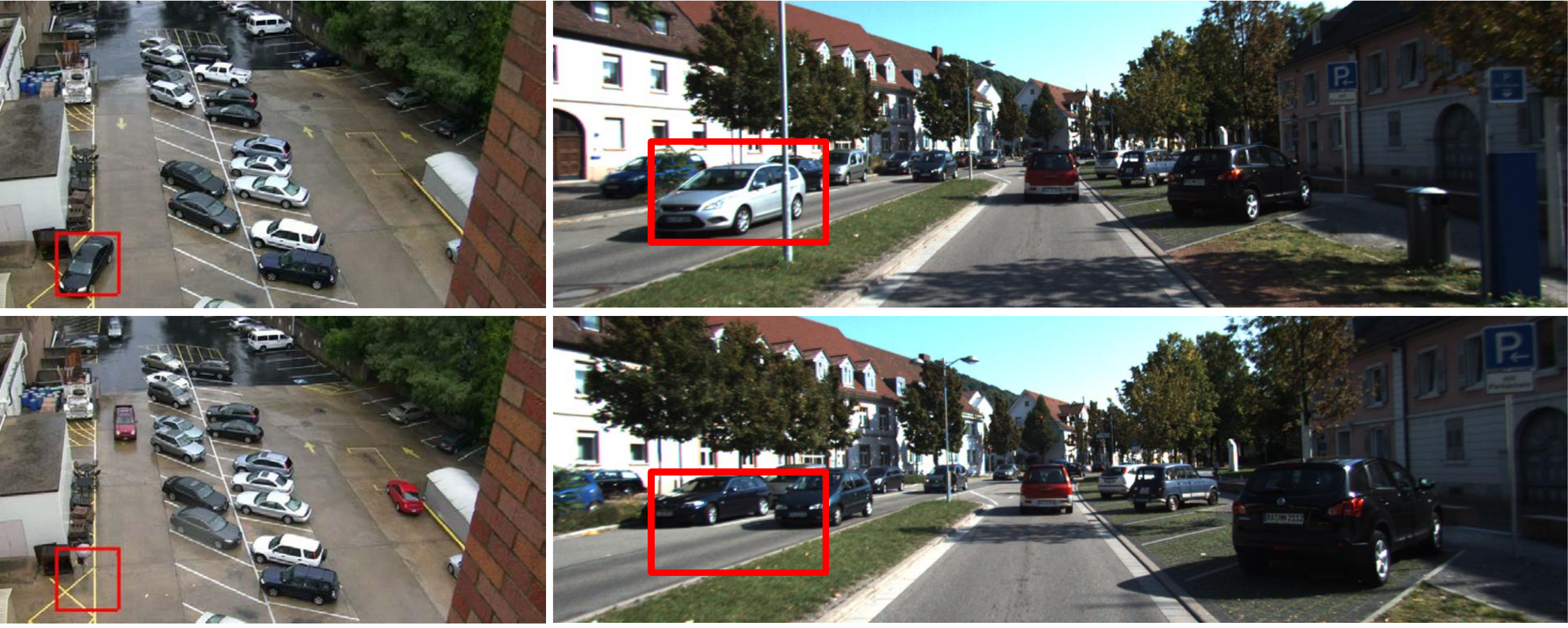}
\caption{Sparsely labeled positives (as shown in the top row) are used to train Exemplar detectors~\cite{tomasz-esvm}. Since we do not assume exhaustive labeling of each instance in the image, we cannot sample negative data from around the input boxes. When these detectors (trained without domain negatives) are used, they may learn background features (like the co-occuring yellow stripes or road divider) and give high confidence false positives (bottom row). We address this issue by exploiting decorrelated errors (Section~\ref{sec:outlierDetails})}

\vspace{-0.2in}
\label{fig:static-back-fp}
\end{figure}
\vspace{-0.1in}
\section{Approach Overview} 
\label{sec:overview}
\vspace{-0.1in}
There are two ways to detect objects in videos -- either using detection in individual frames or tracking across frames.
In a semi-supervised framework, detection plays the role of constraining the system
by providing an appearance prior for the object, while tracking generalizes by
providing newer views of the object. So one could imagine a detection and tracking combination, in which one tracks from confident detections and then updates the detector using the tracked samples. However, as we show in our
experiments (Section~\ref{sec:experiments}), such a na\"{i}ve combination does not impose enough constraints for SSL. In contrast, our approach builds
on top of this basic combination of detection and tracking to correct their mistakes.

Our algorithm starts with a few sparsely annotated video frames
($\mathcal{L}$) and iteratively discovers new instances in the large unlabeled
set of videos ($\mathcal{U}$). Simply put, we first train detectors on
annotated objects, followed by detection on input videos. We determine good
detections (removing confident false positives) which serve as starting points for
short-term tracking. The short-term tracking aims to label new and unseen examples reliably. Amongst these newly labeled examples, we identify good and diverse examples which are used to update the detector without re-training from scratch. We iteratively repeat this fusion of tracking and detection to label new examples. We now describe our algorithm (illustrated in Figure~\ref{fig:teaser}).

\par \noindent \textbf{Sparse Annotations (lack of explicit negatives):} We
start with a few sparsely annotated frames in a random subset of $\mathcal{U}$. Sparse labeling implies that unlike other
approaches~\cite{tld}, we do not assume exhaustively annotated input, and thus cannot sample negatives from the vicinity of labeled
positives. We use random images from the internet as negative data for training object detectors on these sparse labels~\cite{abhinav-data-driven}. We use these
detectors to detect objects on a \emph{subset of the video}, e.g., every 30 frames.
Training on a few positives without domain negatives can result in high confidence false positives as shown in Figure~\ref{fig:static-back-fp}. Removing such false positives is important because if we track them, we
will add more bad training examples, thus degrading the detector's performance over iterations.

\par \noindent \textbf{Temporally consistent detections:} We first remove
detections that are temporally inconsistent using a smoothness prior on the
motion of detections.

\par \noindent \textbf{Decorrelated errors:} To remove high confidence false
positives (see Figure~\ref{fig:static-back-fp}), we rely on the principle of \emph{decorrelated errors} (similar to
\emph{multi-view} SSL~\cite{coreg-multiview,rosenberg}). The intuition is that the
detector makes mistakes that are related to its feature
representation~\cite{hoggles}, and a different feature representation would
lead to different errors. Thus, if the errors in different feature spaces are
decorrelated, one can correct them and remove false positives. This gives us a filtered set of detections.

\par \noindent \textbf{Reliable tracking:}
We track these filtered detections to label new examples. Our final goal is not to track the object over a long period. Instead, our goal is to track reliably and label new and hopefully diverse examples for the object detector. To get such reliable tracks we design a conservative \emph{short-term tracking} algorithm that identifies \emph{tracking failures}. Traditional tracking-by-detection approaches~\cite{hamed-deva-tracking,geiger-kitti-2} rely heavily on the detection prior to identify tracking failures. In contrast, the goal of our tracking is to improve the (weak) detector itself. Thus, heavily relying on input from the detector defeats the purpose of using tracking in our case.

\par \noindent \textbf{Selection of diverse positives for updating the
detector:} The reliable tracklets give us a large set of automatically labeled
boxes which we use to update our detector. Previous work~\cite{prest-video}
temporally subsamples boxes from videos, treating each box with equal
importance. However, since these boxes come from videos, a large number of
them are redundant and do not have equal importance for training our detector.
Additionally, the relevance of an example added at the current iteration $i$ depends on whether similar examples were added in earlier iterations. One would ideally want
to train (make an \emph{incremental update}) only on new and diverse examples,
rather than re-train from scratch on thousands of largely redundant boxes. We
address this issue by selection and show a way of training only on diverse, new
boxes. After training detectors on diverse examples, we repeat the SSL process to iteratively label more examples.

\par \noindent \textbf{Stopping criterion of SSL:}  It is desirable to have SSL algorithms which automatically determine when they should stop. We stop our SSL once our selection algorithm indicates that it does not have any good candidates to select.

\vspace{-0.05in}
\section{Approach Details}
\label{sec:details}
\vspace{-0.1in}
We start with a small sparsely labeled set $\mathcal{L}_0$ of
bounding boxes and unlabeled input videos $\mathcal{U}$. At each iteration
$i$, using models trained on $\mathcal{L}_{i-1}$, we want to label new boxes
in the input videos $\mathcal{U}$, add them to our labeled set
$\mathcal{L} =\mathcal{L} \cup \mathcal{L}_{i}$, and iteratively repeat this procedure.
\vspace{-0.05in}
\subsection{Detections with decorrelated errors}
\label{sec:outlierDetails}
\vspace{-0.1in}
We train object detectors on our initial set of examples using random
images from Flickr as negatives~\cite{abhinav-data-driven}. We detect on a uniformly
sampled subset of the video frames and remove the temporally inconsistent detections.

Since the negative data for training the detectors comes from a different domain
than the positives, we still get consistent false positive detections because
the detector learns the wrong concept (see Figures~\ref{fig:static-back-fp} and ~\ref{fig:qualAbl}). To remove such false positives, we perform outlier removal in a feature space different from that of the detector. The intuition is that the errors made by
learning methods are correlated with their underlying feature representation,
and thus using decorrelated feature spaces might help correct them.

For outlier removal, we use unconstrained Least Squares Importance Fitting (uLSIF)~\cite{uLSIF}. uLSIF uses definite inliers (our labeled set $\mathcal{L}_{i-1}$) to identify outliers in unknown data. It scales well with the size of the input and can be computed in closed form. These final filtered detections serve as starting points for reliable short term tracking.
\subsection{Reliable Tracking}
\vspace{-0.05in}
We formulate a scalable tracking procedure that effectively capitalizes
on priors available from detection, color/texture consistency, objectness~\cite{objectness,sel-search} and optical flow. More importantly, our tracking procedure is very good at identifying its own failures. This property is vital in our semi-supervised
framework since any tracking failure will add wrong examples to the labeled set
leading to quick semantic drift (see Figure~\ref{fig:qualAbl}). The short-term tracking produces a set of labeled examples $\mathcal{L}_i$.

Our single object tracking computes sparse optical flow using Pyramidal Lucas
Kanade~\cite{bouget-lk} on Harris feature points. Since we start with a small set of labeled examples, and do not perform expensive per-frame detection, our detection prior is
weak. To prevent tracking drift in such cases we incorporate color/texture consistency
by using object proposal bounding boxes~\cite{sel-search} obtained from a
region around the tracked box. We address two failure modes of tracking: 
\par \noindent \textbf{Drift due to spurious motion:} This occurs while computing optical flow
on feature points which are not on the object, e.g., points on a moving background or occlusion. To correct this, we first divide each tracked box into four quadrants and compute the mean flow in each quadrant. We weigh points in each quadrant by their agreement with the flow in the other quadrants. The final dominant motion direction for the box is the weighted mean of the flow for each point. This simple and efficient scheme helps correct the different motion of feature points not on the object.
\par \noindent \textbf{Drift due to appearance change:} This is incorporated by object
detection boxes and object proposal bounding boxes in the trellis graph formulation described below.

We formulate the tracking problem as finding the min-cost path in a graph
$\mathcal{G}$. At each frame we incorporate priors in terms of
bounding boxes, i.e., detection bounding boxes, tracked boxes and
object proposal bounding boxes. These boxes are the nodes in our graph and we
connect nodes in consecutive frames with edges forming a trellis graph. The edge weights are a linear combination
of the difference in dominant motions of the boxes (described above), spatial proximity and area change. Tracking through this trellis graph $\mathcal{G}$ is the equivalent of finding the single min-cost path, and is efficiently computed using Dynamic-Programming~\cite{hamed-deva-tracking,k-shortest}. As post-processing, we cut the path as soon as the total cost exceeds a set threshold.

\vspace{-0.05in}
\subsection{Selection algorithm:} 
\label{sec:selectionDetails}
\vspace{-0.05in}
After we label thousands of boxes $\mathcal{L}_i$ for the current iteration, we
use them for improving our object detectors. Since video data is highly redundant, we label few diverse examples and many redundant ones.
Training a category detector on these thousands of (redundant) boxes, \emph{from scratch}, in
every iteration is suboptimal. We prefer an \emph{incremental training} approach that makes
incremental updates to the detector, i.e., trains only on \emph{newly added and diverse} examples rather than everything. This is especially important to prevent drift because even if we label thousands of wrong but redundant boxes, our method picks only a few of them. We find the exemplar detectors~\cite{tomasz-esvm,bharath-whitening} suitable for incremental learning as they are trained per bounding box.

For each labeled bounding box in $\mathcal{L}_i$, we compute a detection
signature~\cite{object-bank, ishan-wacv,yuxiong} using our exemplar detectors. Boxes where our
current set of detectors do not give a high response correspond to examples
which are not explained well by the existing set of detectors. Training on these
boxes increases the coverage of our detectors. Thus, we compute similarity in
this detection signature space to greedily select a set of boxes that are
neither similar to our current detectors, nor amongst themselves.

More formally, let $\mathcal{L}_{i} = \{l_1, \ldots, l_k\}$ be the set of labeled boxes at iteration $i$ and $\mathcal{E} = \cup_{j=0}^{i-1} \mathcal{D}_j$ the set of boxes $b_n$ associated with the exemplar detectors from all previous iterations ($0$ to $i-1$). We compute the $|\mathcal{L}_{i}| \times |\mathcal{E}|$ detector response matrix $R$. The entry $R(m,n)$ indicates the response of the detector associated with box $b_n$ on box $l_m$ (also called the detection signature~\cite{object-bank}). We row-normalize the matrix $R$, and denote the detection signature of box $l_m$ by its row $R(m)$. We initialize the set $\mathcal{D}_{i} \subset \mathcal{L}_{i}$ with all the boxes $l_j$ which have a low response, i.e., none of the detectors from the previous iterations are confident about detecting these boxes (ESVM score of $<-0.8$ and IOU $<0.4$).

We iteratively grow the set $\mathcal{D}_{i}$ by adding detectors which minimize the following objective criterion
\begin{align}
 t^* &= \underset{l_t \in \mathcal{L}_{i}}{\arg\!\max} \sum_{b_p \in \mathcal{D}_i, p\neq t} \mathrm{JSD}( R(t) || R(p) ) \label{eqn:kldiv-selection} \\
 D_{i} &= D_{i} \cup \{ l_{t^*} \} \nonumber
\end{align}

This objective function favors boxes that are diverse with respect to the existing detectors
using the Jensen-Shannon Divergence (JSD) of the detection signatures. At each
iteration, we limit the number of boxes selected to $10$.
When this selection approach is unable to find new boxes, we conclude that we
have reached the saturation point of our SSL. This serves as our \emph{stopping criterion.}
 \vspace{-0.05in}

\section{Experiments}
\label{sec:experiments}
 \vspace{-0.05in}
Our algorithm has a fair number of components interacting with each other across iterations. It is difficult to characterize the importance of each component by using the whole system at once.
For such component-wise characterization, we divide our experiments in two sets.
Our first set of ablative experiments is on a small subset of videos from VIRAT~\cite{virat}. In the second set of experiments, we demonstrate the scalability of our approach to a million frames from~\cite{virat}. We also show the generalization of our method to a different dataset (KITTI~\cite{geiger-kitti-1}). In both these cases we evaluate the automatically labeled data in terms of quality, coverage (recall), diversity and relevance to training an object detector. We now describe the experimental setup which is common across all our experiments.

\begin{figure}[!t]
        \centering
        \begin{subfigure}[b]{0.21\textwidth}
                \includegraphics[width=\textwidth]{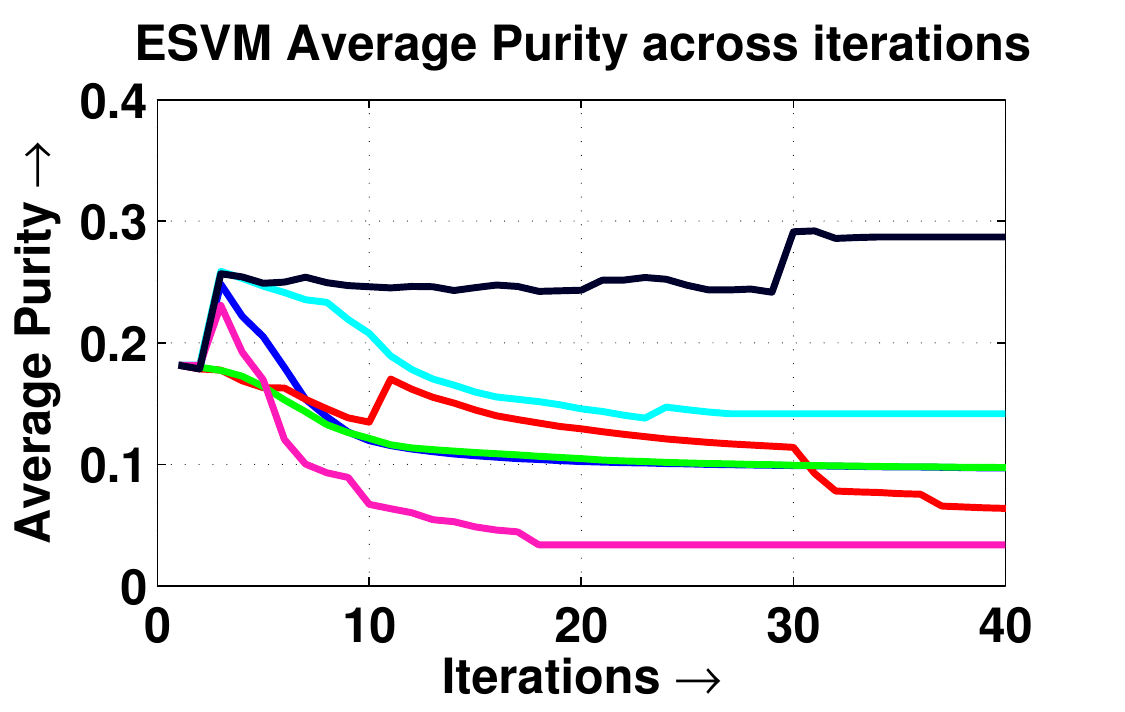}
                \label{fig:tiger}
        \end{subfigure} \hspace{-0.15in}
        \begin{subfigure}[b]{0.21\textwidth}
                \includegraphics[width=\textwidth]{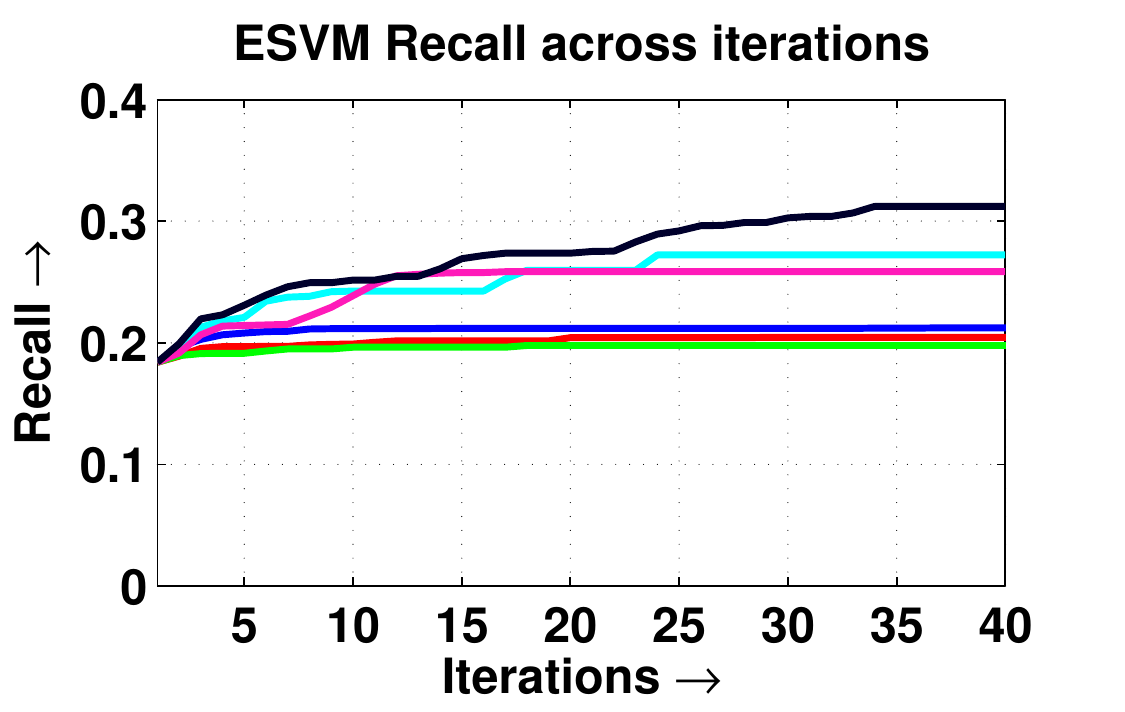}
                \label{fig:mouse}
        \end{subfigure} \hspace{-0.15in}
       \begin{subfigure}[b]{0.06\textwidth}
       \centering
                \includegraphics[width=\textwidth]{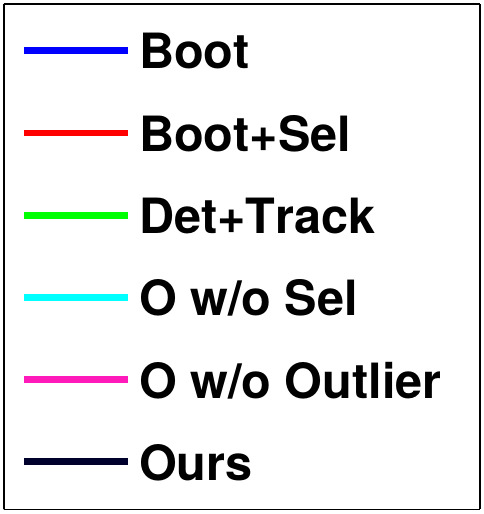}
			\begin{minipage}{0.01in}
            		\vspace{0.75in}
	         \end{minipage}                
                \label{fig:mouse2}
        \end{subfigure}        
        \vspace{-0.6in}
	\caption{We measure the detection performance of the ESVMs from each ablation method on our test set by computing (left) Average Purity and (right) Recall.}	
\vspace{-0.2in}
  \label{fig:abl-graphs}    
\end{figure}

\par \noindent \textbf{Datasets:}
Due to limited availability of large video datasets with bounding box annotations, we picked car as our object of interest, and two video datasets with a large number of cars and related objects like trucks, vans etc. We chose the VIRAT 2.0 Ground~\cite{virat} dataset for its large number of frames, and \emph{sparsely} annotated bounding boxes over all the frames. This dataset consists of long hours of surveillance videos (static camera) of roads and parking lots. It has 329 videos ($\sim$1 million frames, and $\sim$6.5 million annotated bounding boxes (partial ground truth) of cars. We also evaluate on videos from the KITTI~\cite{geiger-kitti-1} dataset which were collected by a camera mounted on a moving car. We use the set 37 videos ($\sim$12,500 frames) which have partial ground truth boxes ($\sim$41,000 boxes, small cars are not annotated).

\par \noindent \textbf{Dataset characteristics:} We chose these datasets with very different characteristics (motion, size of object etc.) to test the generalization of our method. The VIRAT and KITTI datasets both consist of outdoor scene videos with a static and moving camera respectively. The VIRAT dataset captures surveillance videos of multiple cars in parking lots. The cars in this dataset are small compared to the frame size, tightly packed together and viewed from a distance (thus no drastic perspective effects). The KITTI dataset on the other hand, consists of videos taken by a vehicle mounted camera. It has high motion, large objects, and perspective effects. Figure~\ref{fig:static-back-fp} shows examples from both the datasets demonstrating their visual differences.

\par \noindent \textbf{Detectors:} We use the Exemplar-SVM (ESVM)~\cite{tomasz-esvm} detectors with 5000 random images from Flickr as negatives~\cite{abhinav-data-driven}. Since per frame detection is expensive, we detect once every $30^\text{th}$ frame for VIRAT, and every $10^\text{th}$ frame for KITTI. We threshold detections at SVM score of $-0.75$.

\par \noindent \textbf{Multiple feature spaces for false positive removal:} We use the uLSIF~\cite{uLSIF} algorithm on Pyramidal HOG (PHOG)~\cite{phog} and color histogram (LAB color with $32\times16\times16$ bins) features computed on a resized box of $200 \times 300$ px. We set the kernel bandwidth for uLSIF by computing the $75^\text{th}$ percentile distance between random pairs of points.

\par \noindent \textbf{Object proposal windows}: We obtain 2000 windows per image using selective search~\cite{sel-search}.
\begin{table*}
    \centering
\small{    
\caption{Comparison of our method with baselines as explained in Section~\ref{sec:exp-Abl}. We train an LSVM~\cite{pedro-dpm} on all the automatically labeled data and compute its detection performance on a held-out, fully annotated test set (AP for IOU 0.5).}
\label{tbl:lsvms}
\vspace{-0.1in}
\begin{tabular}[b]{@{}c@{}p{1em}@{}cccccc@{}p{1em}@{}ccc@{}}\toprule
& & \multicolumn{6}{c}{Automatic Labeling (LSVM)}  & & \multicolumn{3}{c}{Ground Truth} \\
\cmidrule{3-8}\cmidrule{10-12}
Iteration & & Boot. & Boot.+Sel. & Det+Track & O w/o Sel & O w/o Outlier & \textbf{Ours} & & Pascal LSVM & Pascal DPM & VIRAT LSVM \\
\cmidrule{1-8}\cmidrule{10-12}
10 & & 1.32 & 9.09 & 9.09 & 11.21 & 7.32 & \textbf{15.39} &  & \multirow{2}{*}{20.89} & \multirow{2}{*}{29.56} & \multirow{2}{*}{41.38}\\
30 & & 1.94 & 3.03 & 6.59 & 10.83 & 1.41 & \textbf{17.68}  & &  &  & \\
\bottomrule
\end{tabular}
}
\vspace{-0.1in}
\end{table*}

 \vspace{-0.05in}
\subsection{Ablative Analysis of each constraint}
 \vspace{-0.05in}
\label{sec:exp-Abl}
\begin{figure*}
    \centering
    \includegraphics[width=\textwidth]{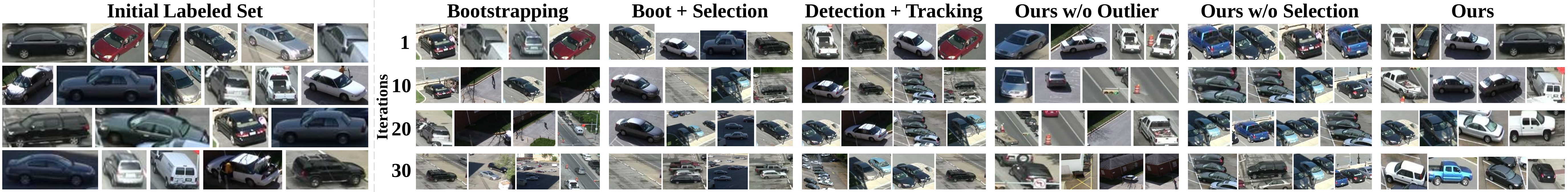}    
\caption{(a) We look at a subset of the bounding boxes used to train ESVMs across iteration. Each row corresponds to an ablation method. The top row shows the randomly chosen initial positive bounding boxes (same for each method). The other methods diverge quickly across iterations, thus showing that constraints are very important for maintaining purity.}
\vspace{-0.2in}
\label{fig:qualAbl}
\end{figure*}
To tease apart the contributions of each component described in Section~\ref{sec:overview}, we design a set of algorithms using only a subset of the components.
\par \noindent \textbf{Bootstrapping (Boot):} In this vanilla form of SSL, we train object detectors on the initial labeled set and perform detection. The confident detections are used as training examples for the next iteration detectors.

\par \noindent \textbf{Bootstrapping with Selection (Boot+Sel):} This algorithm builds upon the bootstrapping approach described above. However, diverse examples are selected (Section~\ref{sec:selectionDetails}) from confident detections to train new detectors.

\par \noindent \textbf{Detection, Tracking and Clustering (Det+Track):} In this algorithm, we use a basic combination of detection and tracking. We start tracking from the confident ESVM detections to label new examples. We then use WHO (or ELDA)~\cite{bharath-whitening} clustering~\cite{ishan-wacv} on these boxes to select training examples for the new detectors. For clustering, we use WHO features on labeled boxes after resizing, followed by $k$-means. We choose the best $k$ in the range $(5,10)$.
\par \noindent \textbf{Ours without outlier (O w/o Outlier):} This setup uses our entire algorithm except outlier removal (Section~\ref{sec:outlierDetails}). It can roughly be thought of as Detection+Tracking+Selection.
\par \noindent \textbf{Ours without selection (O w/o Sel):} This algorithm uses our algorithm except selection (Section~\ref{sec:selectionDetails}). It uses WHO clustering for selection like Det+Track.
\par \noindent \textbf{Ours:} We use our full algorithm as detailed in Sections~\ref{sec:overview}.
\par \noindent \textbf{Ablation dataset:} For these set of experiments we use an input set of 25 videos ($\sim$170,000 frames) and a separate test set of 17 videos ($\sim$105,000 frames) which we fully annotated. All algorithms start with the same sparse labels consisting of only 21 boxes spread across different videos. We run them iteratively till 30 iterations.
\vspace{-0.1in}
\subsubsection{Qualitative Results}
\vspace{-0.1in}
Figure~\ref{fig:qualAbl} shows a random set of boxes labeled by each ablation method (and used to train ESVMs for that method), along with the initial set of 21 positive examples. We notice that as iterations proceed, the labeling quality (especially ``tightness'' of the boxes) for all methods degrades. More importantly, the other methods like Boot, Det+Track etc.\ show semantic drift (Figure~\ref{fig:qualAbl} columns 2-5 at iteration 20). We also notice the importance of selection, as Ours w/o Selection loses good localization ability fairly quickly (Figure~\ref{fig:qualAbl} column 6 at iterations 10-30). We believe methods like~\cite{derek-relo} can be used to further improve the labeling quality.
\vspace{-0.1in}
\subsubsection{ESVM Detection performance}
\vspace{-0.1in}
For the input videos, we cannot measure labeling purity because of partial ground truth (refer to supplementary material for an approximation of purity). Instead, we measure the relevance of labeled boxes to detection. We consider detection performance on the test set as a proxy for good labeling. We test the ESVMs selected and trained by each method across iterations on the held out test set. A good labeling would result in an increased detection performance. Figure~\ref{fig:abl-graphs} shows Average Purity vs.\ Recall across iterations for the various methods on the test set. We use Average Purity, which is same as Average Precision~\cite{pascal-voc-2007} but does not penalize double-detections, since we are more interested in whether the ESVMs are good detectors individually, rather than as an ensemble.
We consider an instance correctly labeled (or pure) if its Intersection-Over-Union (IOU) with any ground-truth instance is greater than 0.3.
Our method shows a higher purity and recall, pointing towards meaningful labeling and selection of the input data. It also shows that every component of our method is crucial for getting good performance. We stop our method at iteration 40 because of our stopping criterion (Section~\ref{sec:selectionDetails}). We got a $2$ point drop in purity from iteration 40 to 45, proving the validity of our stopping criterion. This is important since our algorithm would rather saturate than label noisy examples.
\vspace{-0.1in}
\subsubsection{Training on all the automatically labeled data}
\vspace{-0.1in}
In this section, we evaluate the effectiveness of all our labeled data. 
For each algorithm, we \textbf{train an LSVM}~\cite{pedro-dpm} (only root filters, mixtures and positive latent updates of DPM~\cite{pedro-dpm}) on the data it labeled, and test it on the held-out test set. Since it is computationally expensive to train an LSVM on the thousands of boxes, we subsample the labeled boxes (5000 boxes in total for each method using $k$-means on WHO~\cite{bharath-whitening} features). We sample more boxes from the earlier iterations of each method, because their labeling purity decreases across iterations (Figure~\ref{fig:abl-graphs}). We use the same domain independent negatives~\cite{abhinav-data-driven} for all these LSVMs (left side in Table~\ref{tbl:lsvms}). Table~\ref{tbl:lsvms} shows the detection AP performance of all \textbf{LSVMs} (measured at IOU of 0.5~\cite{pascal-voc-2007}) for the data labeled at iteration 10 and 30. We see that LSVM trained on our labeled data outperforms all other LSVMs. Our performance is close to that of an LSVM trained on the PASCAL VOC 2007 dataset~\cite{pascal-voc-2007}. This validates the high quality of our automatically labeled data.

We also note that the performance of an LSVM trained on the ground truth boxes (VIRAT-LSVM) (5000 boxes from $\sim$1 million ground truth boxes using the same $k$-means as above) achieves twice the performance. The fact that all LSVMs (except the ones from PASCAL) are trained with the same domain-independent negatives, indicates that the lack of domain-negatives is not the major cause of this limitation. This suggests that automatic labeling has limitations compared to human annotations. On further scrutiny of the LSVM trained on our automatically labeled data, we found that the recall saturates after iteration 30. However, the precision was within a few points of VIRAT-LSVM. Since we work with only the confident detections/tracklets in the high precision/low recall regime, this behavior is not unexpected. This is also important since our algorithm would rather saturate than label noisy positives.
\begin{figure*}
        \centering        
        \begin{subfigure}[t]{0.23\textwidth}
                \includegraphics[width=\textwidth]{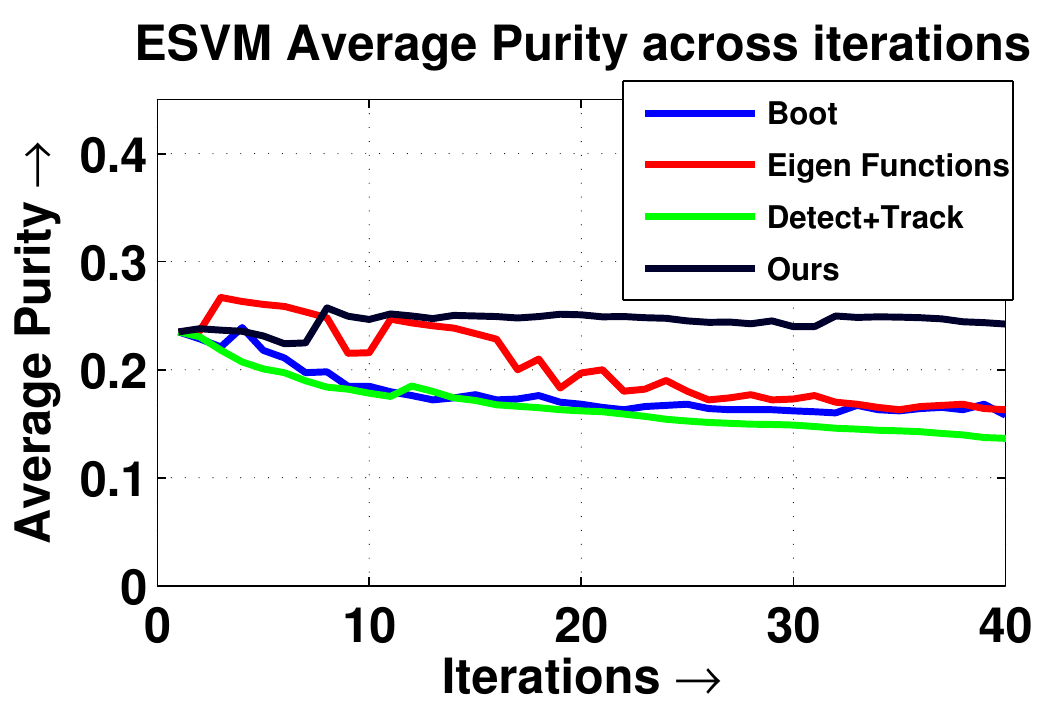}
                \caption{Purity on KITTI}
        \end{subfigure}~%
        \begin{subfigure}[t]{0.23\textwidth}
                \includegraphics[width=\textwidth]{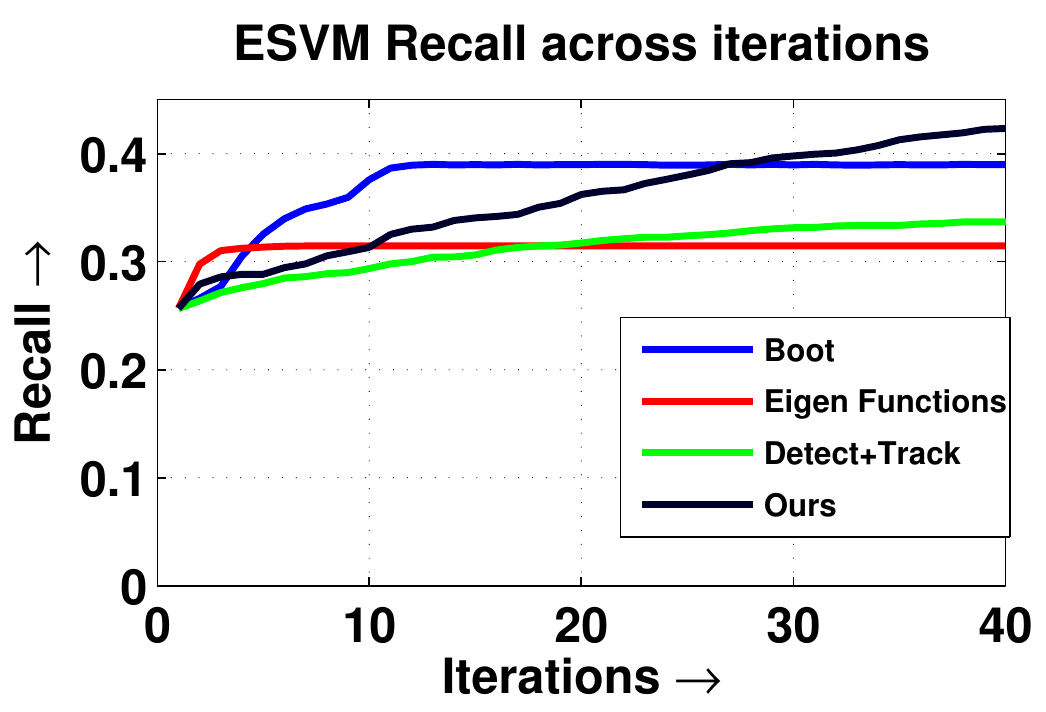}                
                \caption{Recall on KITTI}
        \end{subfigure}~
			\begin{subfigure}[t]{0.23\textwidth}
                \includegraphics[width=\textwidth]{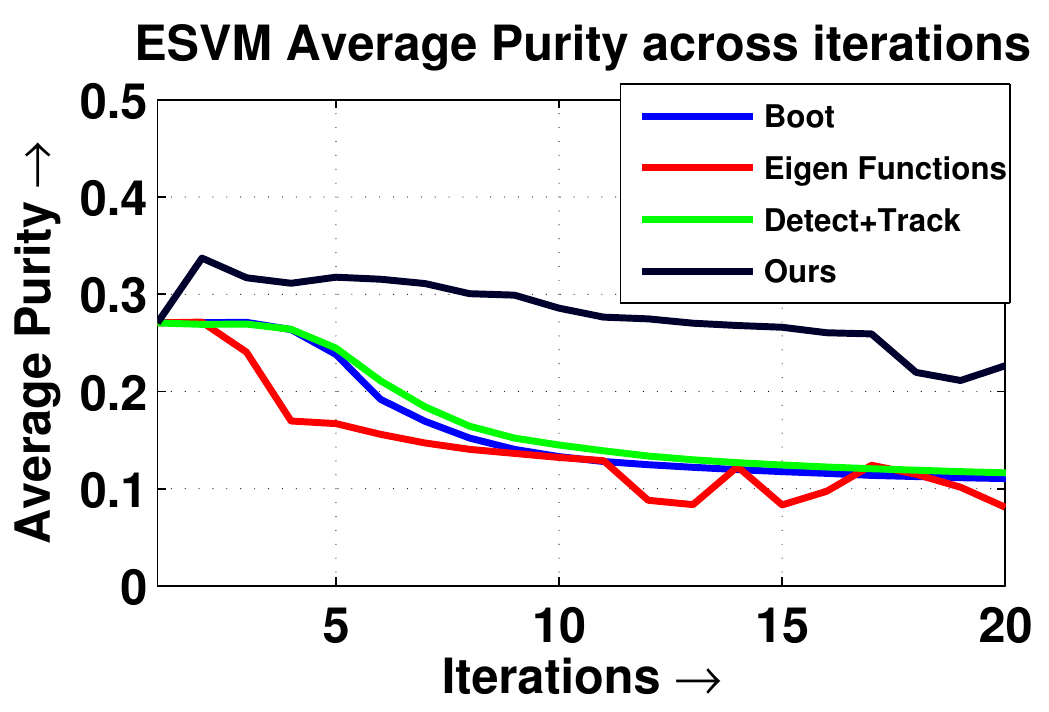}
                \caption{Purity on VIRAT}
        \end{subfigure}~%
        \begin{subfigure}[t]{0.23\textwidth}
                \includegraphics[width=\textwidth]{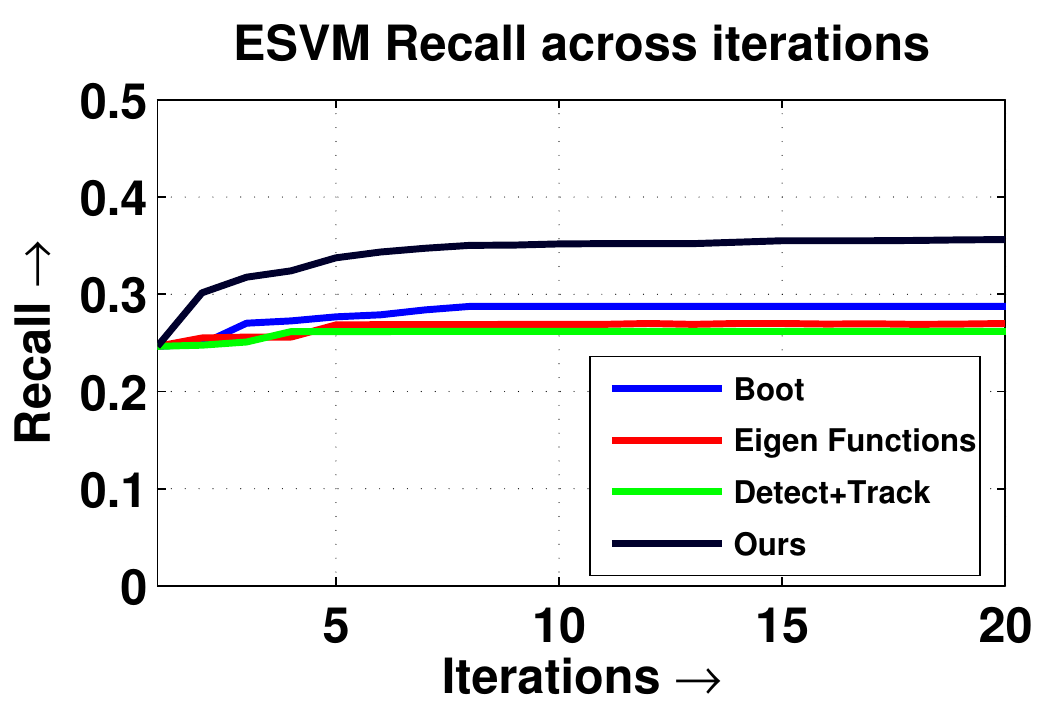}                
                \caption{Recall on VIRAT}
        \end{subfigure}              
         \vspace{-0.05in}
        \caption{We measure the detection performance of the labeled boxes for our large scale experiments. We test the ESVMs trained at each iteration on the held out test set and compute Average Purity and Recall. Our method outperforms the baselines by a significant margin. It maintains purity while substantially increasing recall.}
         \vspace{-0.05in}
        \label{fig:largeScaleGraphs}
\end{figure*}

\begin{figure*}
        \centering        
        \begin{subfigure}[t]{0.16\textwidth}
                \includegraphics[width=\textwidth]{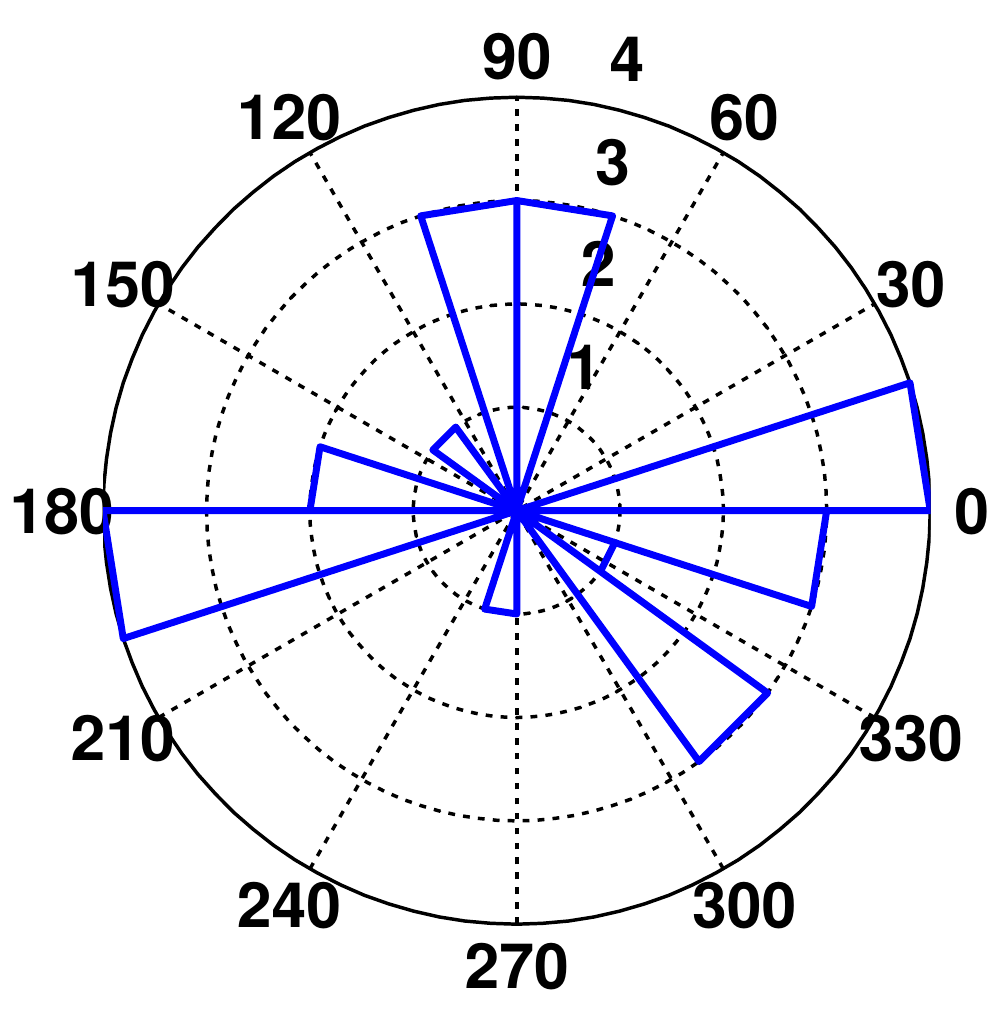}                
                \caption{Initial boxes}
        \end{subfigure}~
        \begin{subfigure}[t]{0.16\textwidth}
                \includegraphics[width=\textwidth]{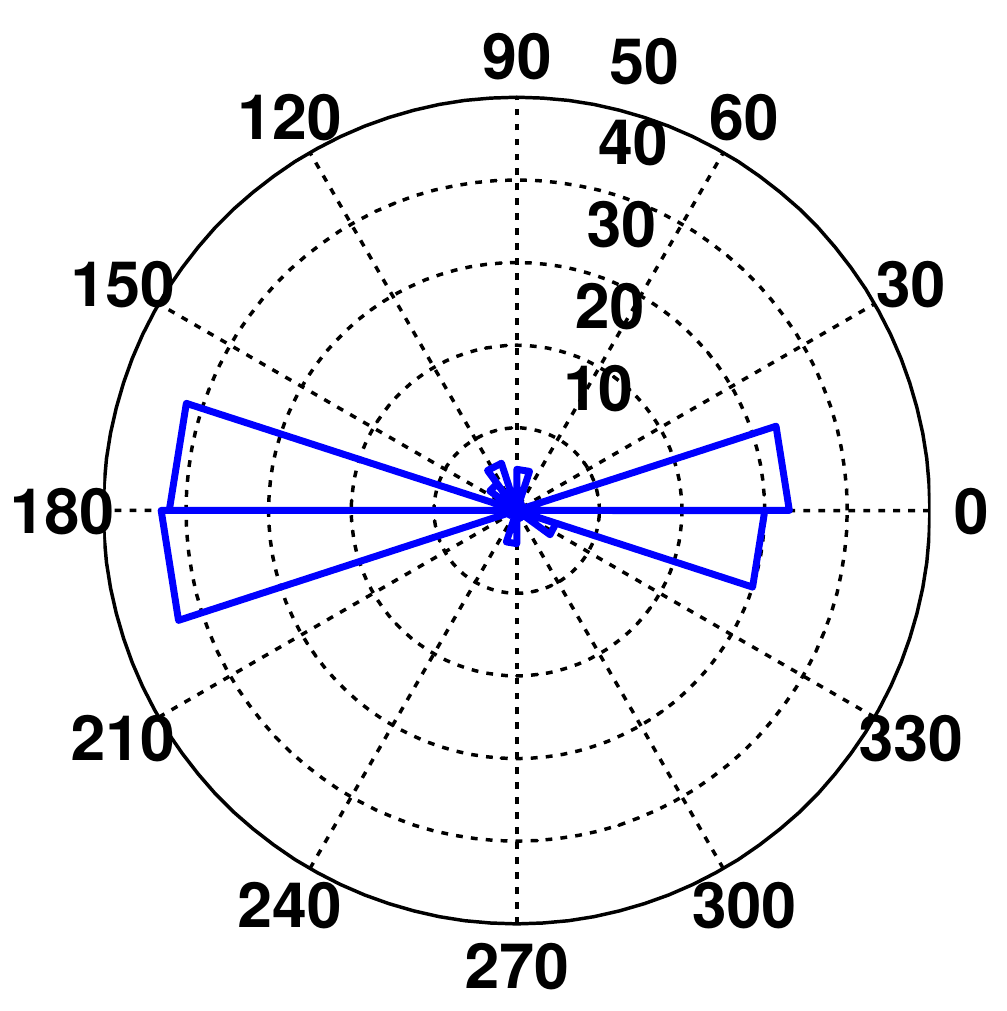}                
                \caption{Bootstrapping}
        \end{subfigure}~          
        \begin{subfigure}[t]{0.16\textwidth}
                \includegraphics[width=\textwidth]{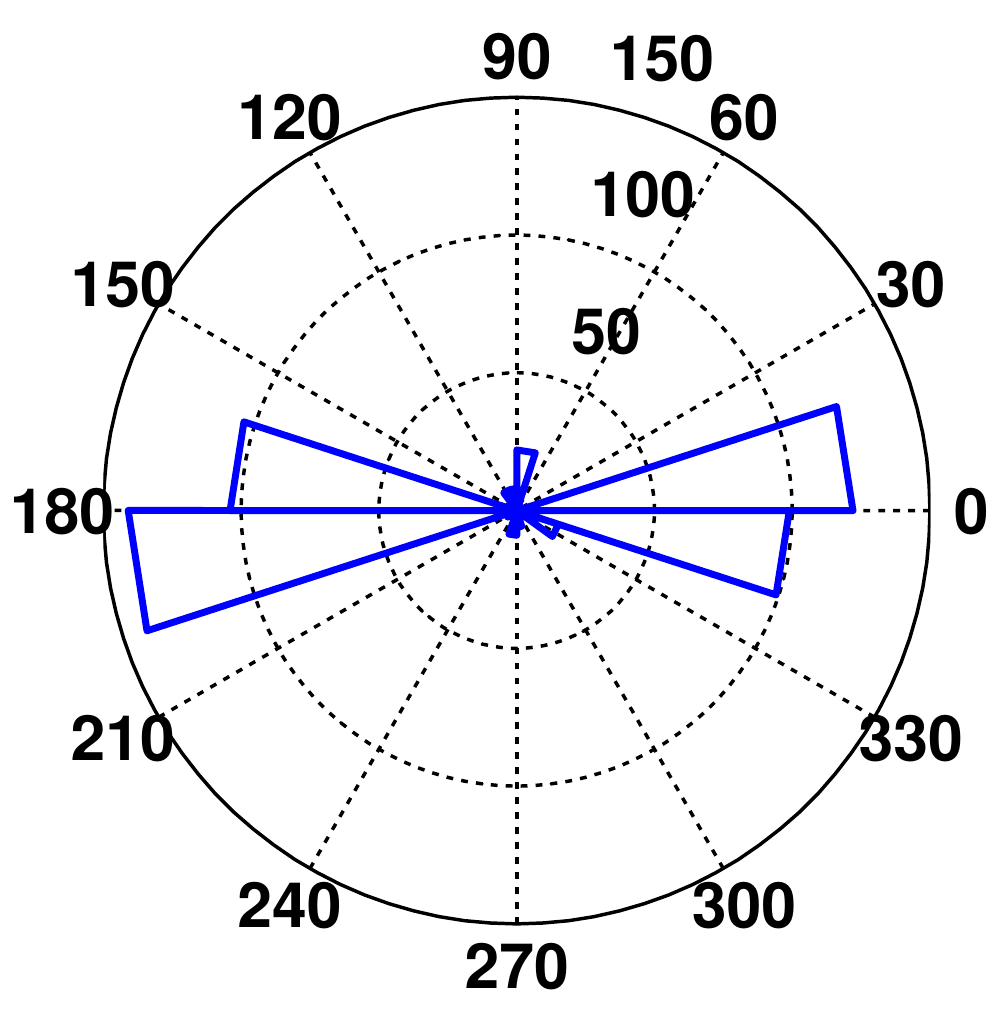}                
                \caption{Eigen Functions}
        \end{subfigure}~                            
        \begin{subfigure}[t]{0.16\textwidth}
                \includegraphics[width=\textwidth]{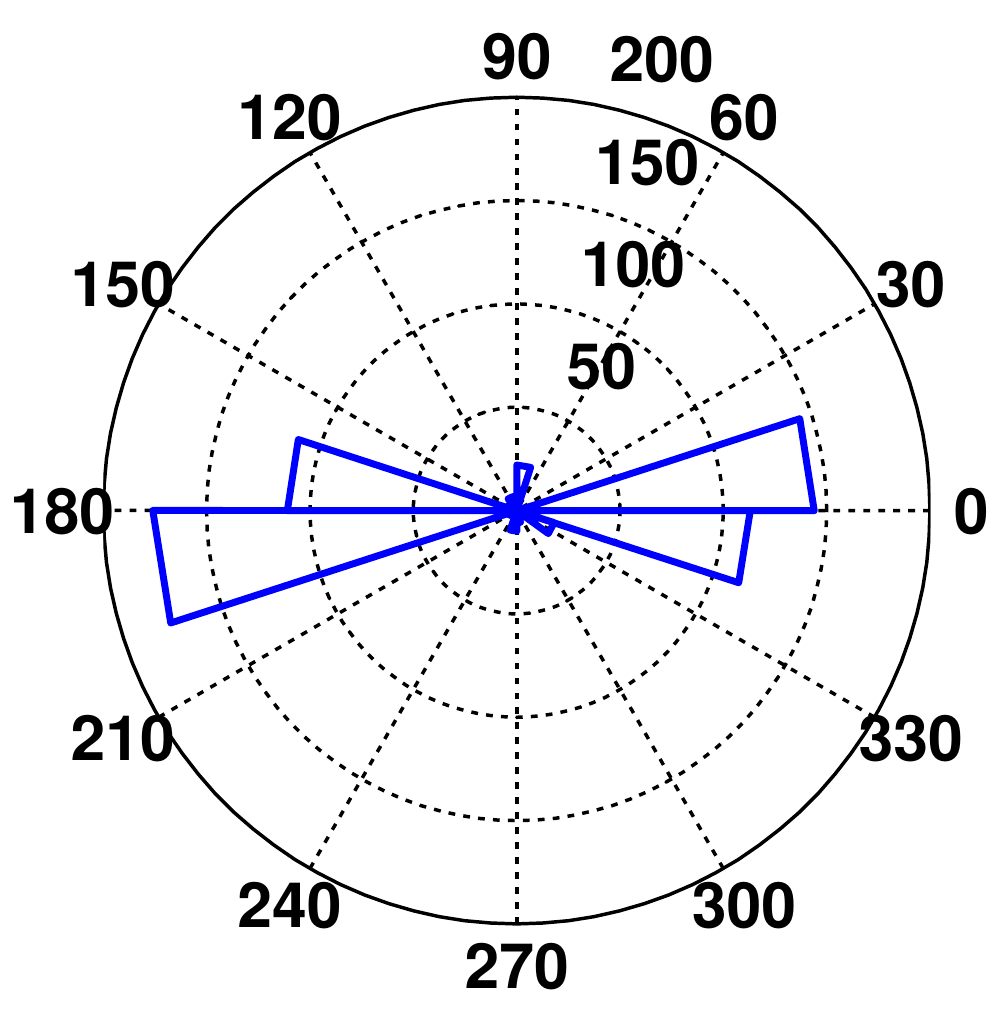}                
                \caption{Detect+Track}
        \end{subfigure}~      
        \begin{subfigure}[t]{0.16\textwidth}
           \includegraphics[width=\textwidth]{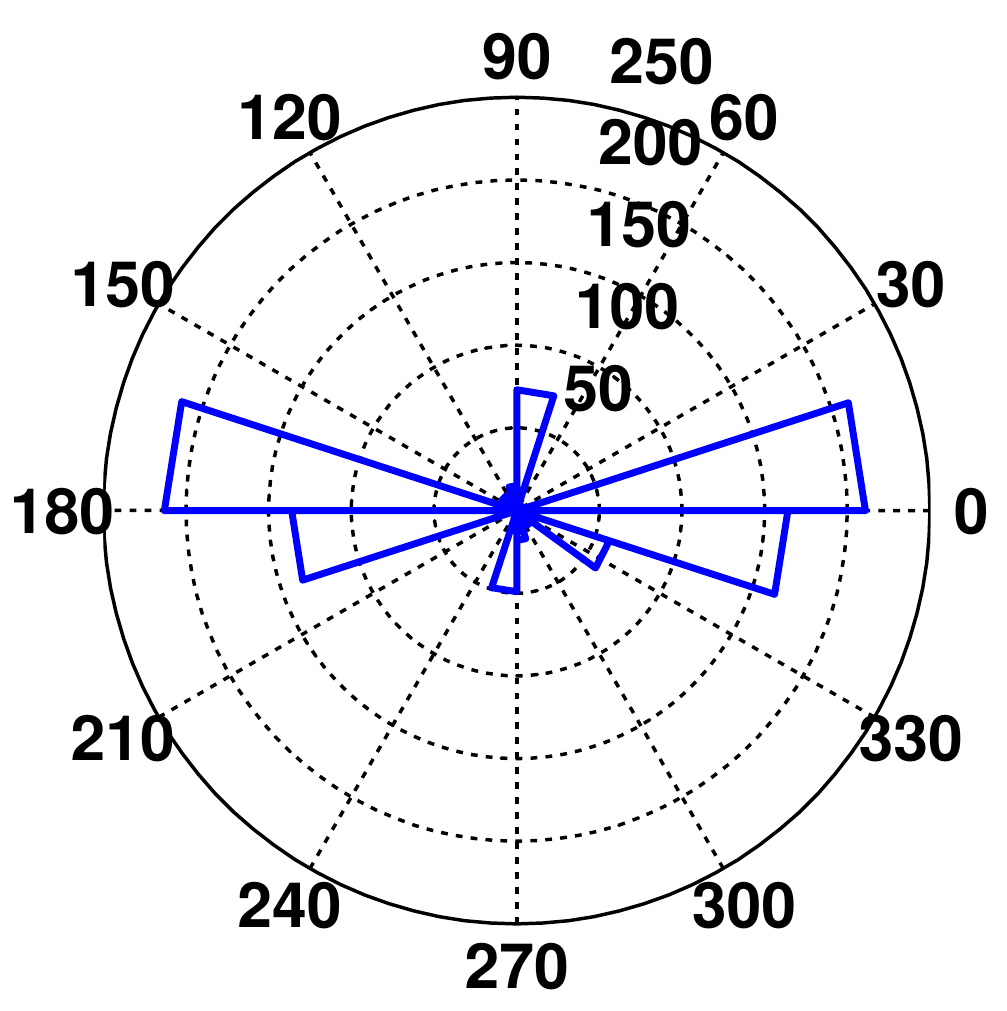}
            \caption{Ours}
        \end{subfigure}~
                \begin{subfigure}[t]{0.16\textwidth}
                \includegraphics[width=\textwidth]{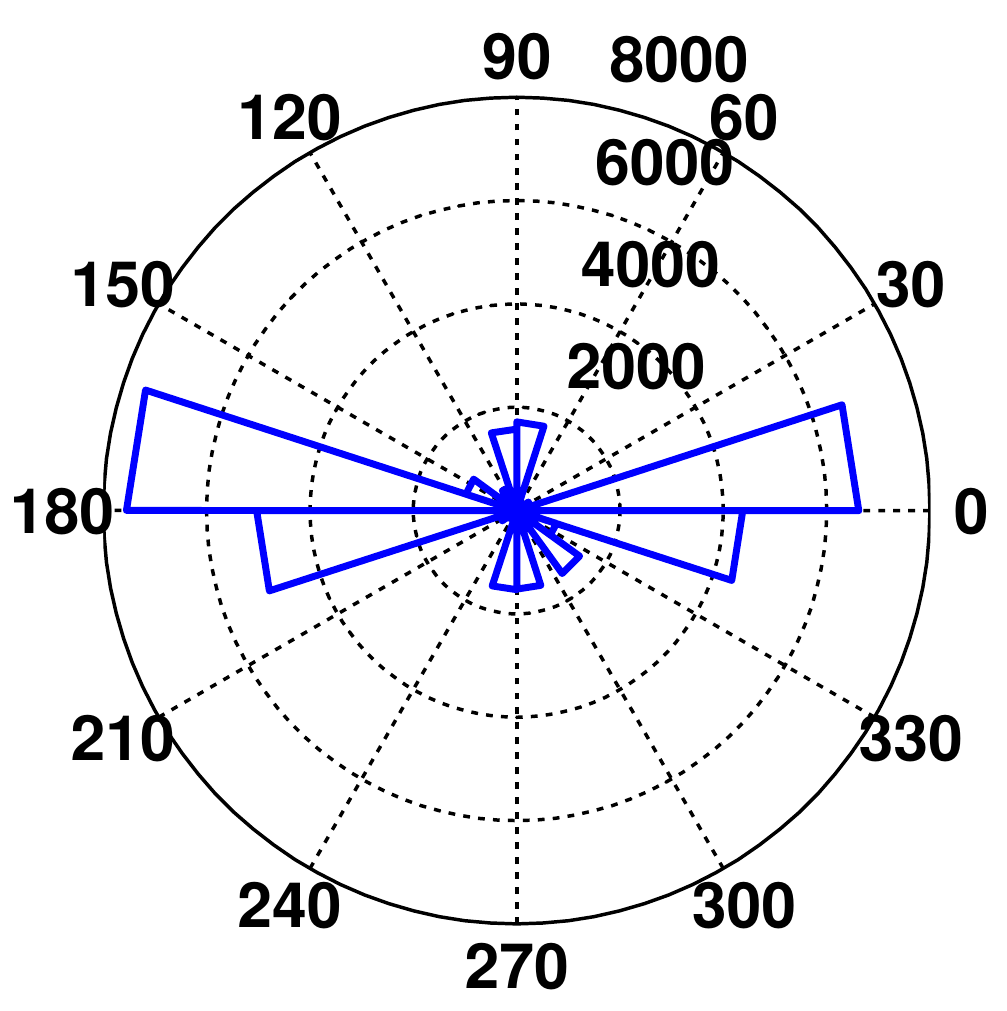}
                \caption{Ground truth}
        \end{subfigure}
       \vspace{-0.1in}
        \caption{Pose variation in automatic labeling of the KITTI dataset. For each algorithm, we plot the 3D pose distribution of all the boxes it labels after 30 iterations. The first and last plots show pose distribution for the initial labeled boxes and all boxes in ground truth respectively. The distribution of boxes labeled by our method is close to the ground truth distribution.}
        \label{fig:poseKitti}
        \vspace{-0.1in}
\end{figure*}

 \begin{figure*}[!tbh]
 \centering
 \includegraphics[width=\textwidth]{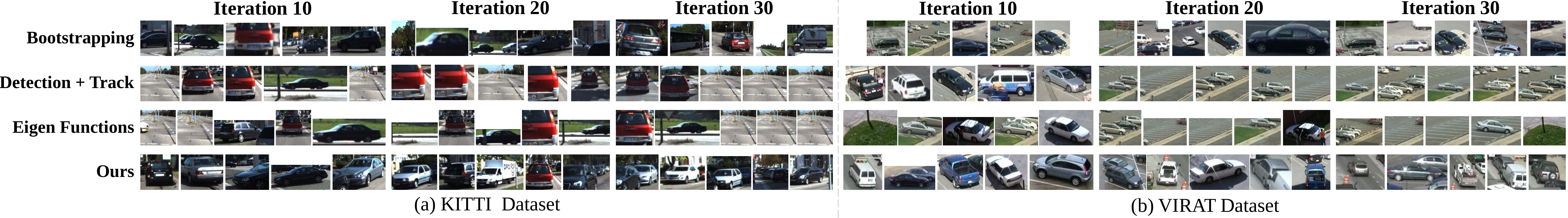}       
  \vspace{-0.2in} 
 \caption{We look at the selected positives for each baseline method across iterations for both KITTI and VIRAT datasets. We notice that the purity of the labeled set drops significantly as iterations proceed. This indicates that constraints specific to video are needed to learn meaningfully over iterations.}
  \vspace{-0.25in}
 \label{fig:largeQual}
 \end{figure*}

\subsection{Large scale experiments}
\label{sec:exp-largeScale}
\vspace{-0.1in}
In this section, we evaluate the scalability of our algorithm to millions of frames and object instances. We also test its generalization on two datasets with widely different characteristics - VIRAT (static camera, small cars tightly packed) and KITTI (moving camera, high motion, large cars with perspective effects). We use the \emph{Boot} and \emph{Ours} methods described in Section~\ref{sec:exp-Abl}. As we described in Section~\ref{sec:relatedWork}, most of the existing approaches make additional assumptions that are not applicable in our setting, or do not have publicly available code. To make a fair comparison against existing methods, we adapt them to our setting.

\par \noindent \textbf{Baseline - Dense detection and association (Detect + Track):} This algorithm is inspired by Geiger et al.~\cite{geiger-kitti-2} which has state-of-the-art results on KITTI. The original algorithm uses per-frame detections, followed by a Kalman filter and data association. We make two changes - 1) To have a comparable detector across methods, we do not use a pre-trained DPM on PASCAL VOC 2007. We substitute it with the ESVMs we have at the current iteration. 2) We do not use a Kalman filter and use data association over a short term (maximum 300 frames). We select positives for the ESVMs by $k$-means on WHO features~\cite{ishan-wacv}.

\par \noindent \textbf{Baseline - Eigen Functions:} We modify the Eigen functions~\cite{fergus-ssl} method which was originally designed for image classification. This method uses distances over manifolds to label unlabeled data. We use basic detection and short term tracking to get a set of bounding boxes and use eigen functions to classify them as positive or background. The boxes from previous iterations on which we trained ESVMs are used as positives and random images from Flickr~\cite{abhinav-data-driven} as negative data. We use color histogram ($32 \times 16 \times 16$ LAB space) and PHOG~\cite{phog} features as input to eigen functions.

\par \noindent \textbf{Datasets:} Our input set consists of 312 videos ($\sim$820,000 frames) from VIRAT. We take a held out test set of 17 videos ($\sim$105,000 frames) which we fully annotated. As input, all algorithms start with the same sparse labels consisting of 43 randomly chosen bounding boxes across different videos.
For the KITTI dataset we use 30 videos ($\sim$10,000 frames) as our input and 7 videos ($\sim$2000 frames) for testing. All algorithms start with the same sparse labeled 25 bounding boxes from different videos.

\par \noindent \textbf{Qualitative Results:}
We first present qualitative results in Figure~\ref{fig:largeQual}. We notice the same trends as we did in the ablation analysis, namely, bounding boxes tend to get less tight across iterations. For the baseline methods, we notice quick divergence as an increasing number of background patches are classified as car.

\par \noindent \textbf{ESVM Detection Performance:}
Following the approach outlined in Section~\ref{sec:exp-Abl}, we compute the detection performance of the ESVMs on the held out test set. This helps us measure the relevance of our labeling to the detection task. Figure~\ref{fig:largeScaleGraphs} shows the results of these experiments. We notice that our method outperforms the baselines on both the metrics (Average Purity and Recall). This reinforces the fact that our constraints help arrest semantic drift.

\par \noindent \textbf{Diversity of labeling:}
The KITTI dataset provides the 3D pose associated with each labeled car. We use this 3D pose as a proxy for estimating the diversity of our labeled set. In this experiment, we compute the pose of the examples labeled by all methods. Figure~\ref{fig:poseKitti} demonstrates that our labeling procedure covers a diverse range of poses as opposed to baseline methods. The pose distribution of our labeling is closer to the ground truth distribution, while that of the baselines prefers the more ``popular'' poses, i.e., front/back of cars. Combined with the results of Figure~\ref{fig:largeScaleGraphs}, this points towards a diverse, and high quality labeling of the data.

 \vspace{-0.1in}
\section{Conclusions}
 \vspace{-0.1in}
We introduce a semi-supervised learning technique for training object detectors from videos. Our technique addresses the detection of multiple objects without assuming exhaustive labeling of object instances in any input frame. In addition, we introduce constraints like decorrelated errors, reliable tracking and diverse selection which are effective in arresting semantic drift. Our experiments show that such an SSL approach can start with a handful of labeled examples and label hundreds of thousands of new examples which also improve object detectors.

\smallskip
\small{
\noindent \textbf{Acknowledgments:}
The authors thank Robert Collins and the reviewers for many helpful comments. This project was supported by NSF Grant IIS1065336 and a Google Faculty Research Award. IM was also supported by the Siebel Scholarship, and AS was supported by the Microsoft Research PhD Fellowship.}

{\small
\bibliographystyle{ieee}
\bibliography{myRefs}
}

\end{document}